\documentclass[12pt]{article}
\usepackage{fancyvrb}
\usepackage{authblk}
\usepackage{titlesec}
\usepackage{geometry}
\author{Anatole Callies\thanks{Inato} , Quentin Bodinier$^*$, Philippe Ravaud \thanks{Center for Research in Epidemiology and Statistics (CRESS), Université Paris Cité and Université Sorbonne Paris Nord, INSERM, INRAE, F-75004 Paris, France} \thanks{Centre d'epidémiologie clinique, AP-HP, Hôpital Hôtel Dieu, F-75004 Paris, France} $^*$ and Kourosh Davarpanah$^*$}

\usepackage{multirow}
\usepackage{subcaption}
\usepackage{graphicx} 
\graphicspath{{Images/}}
\usepackage[
style=authoryear,
backend=biber,
hyperref=true,
backref=true
]{biblatex}
\usepackage[colorlinks=true, citecolor=blue]{hyperref} 

\providecommand{\keywords}[1]
{
  \small	
  \textbf{\textit{Keywords---}} #1
}

\title{Real-world validation of a multimodal LLM-powered pipeline for High-Accuracy Clinical Trial Patient Matching leveraging EHR data}

\date{\vspace{-1em}}

\begin{document}

\maketitle

\begin{abstract}
\textbf{Background}: Patient recruitment in clinical trials is hindered by complex eligibility criteria and labor-intensive chart reviews. Prior research using text-only models have struggled to address this problem in a reliable and scalable way due to (1) limited reasoning capabilities, (2) information loss from converting visual records to text, and (3) lack of a generic EHR integration to extract patient data.

\textbf{Methods}: We introduce a broadly applicable, integration-free, LLM-powered pipeline that automates patient–trial matching using unprocessed documents extracted from EHRs. Our approach leverages (1) the new reasoning-LLM paradigm, enabling the assessment of even the most complex criteria, (2) visual capabilities of latest LLMs to interpret medical records without lossy image-to-text conversions, and (3) multimodal embeddings for efficient medical record search. The pipeline was validated on the n2c2 2018 cohort selection dataset (288 diabetic patients) and a real-world dataset composed of 485 patients from 30 different sites matched against 36 diverse trials.

\textbf{Results}: On the n2c2 dataset, our method achieved a new state-of-the-art criterion-level accuracy of 93\%. In real-world trials, the pipeline yielded an accuracy of 87\%, undermined by the difficulty to replicate human decision-making when medical records lack sufficient information. Nevertheless, users were able to review overall eligibility in under 9 minutes per patient on average, representing an 80\% improvement over traditional manual chart reviews.

\textbf{Conclusion}: This pipeline demonstrates robust performance in clinical trial patient matching without requiring custom integration with site systems or trial-specific tailoring, thereby enabling scalable deployment across sites seeking to leverage AI for patient matching.
\end{abstract}


\newpage

\section{Introduction}
Patient recruitment remains a critical bottleneck in clinical development pipelines, with enrollment challenges alone accounting for 30\% of phase III trial terminations and substantially prolonging drug development timelines (\cite{shah2021accelerating}). At the core of this problem is the laborious task of identifying suitable patients — a step often complicated by increasingly stringent requirements for trial eligibility. Over the last two decades, the median number of eligibility criteria has surged by 58\%, rising from 31 in 2001--2005 to 49 in 2016--2020 (\cite{shah2021accelerating}). This added complexity is particularly prevalent in fields such as oncology, neurology, and infectious diseases, where the criteria counts can exceed 60 (\cite{shah2021accelerating}), placing even greater demands on Clinical Research Coordinators (abbreviated as CRCs in the rest of the document). As a result, the pre-screening phase alone can consume up to 60\% of total patient recruitment effort (\cite{penberthy2012effort}). When done manually, each patient pre-screening takes on average 50 minutes in phase III trials. Because as many as 88\% of them ultimately fail to meet the criteria, it takes on average more than seven hours to find one eligible patient for Phase III trials (\cite{penberthy2012effort}).

In the \textbf{Related Work} section, we review previous efforts to address this challenge, detailing their approaches and the performance they achieved. In the \textbf{Methods} section, we present our pipeline, and how it overcomes those limitations by leveraging several recent innovations. The \textbf{Results} section presents tangible evidence of our pipeline’s effectiveness, exceeding state-of-the-art performance on both a public benchmark (the n2c2 2018 cohort selection dataset) and our real-world dataset built from in-app user feedbacks. We also demonstrate how it significantly decreases pre-screening time. Finally, the \textbf{Discussion} section synthesizes our findings and suggests future enhancements around uncertainty calibration and patient-level recommendations.

\section{Related work}

\subsection{Previous research}
Prior research has tackled this challenge through two main approaches. First, a "patient-centric" approach to find the relevant trials for a given patient. And, on the contrary, a "trial-centric" approach to find relevant patients for a given trial. In this study, we focus on the latter "trial-centric" task and adapt previous approaches for a better real-world applicability.
Many previous studies have proven the theoretical feasibility and effectiveness of LLM-based eligibility assessments. To start with, \cite{wong2023scalingclinicaltrialmatching} introduced a framework leveraging LLMs to convert eligibility criteria into queryable expressions and match them against electronic health records (EHRs). Their approach demonstrated a patient-level recall varying between 67.3\% and 76.8\%. Though, their focus was solely on the challenging area of oncology trials, making it difficult to generalize to other therapeutic areas. Similarly, \cite{hamer2023improvingpatientprescreeningclinical} explored the use of LLMs, and particularly OpenAI’s InstructGPT (pre-ChatGPT chat assistant), to perform patient-trial matching on a synthetic patient dataset. They showed that while the AI standalone performance was sub-par (low criterion-level accuracy of 72\% resulting in a patient-level recall of 50\% only), it was still possible to use it as a copilot and thus achieve perfect patient-level recall and acceptable precision (71\%) while reducing the number of criteria to be checked by experts by 90\%. Then, \cite{Jin_2024} introduced TrialGPT, a three-stage approach (retrieval, criterion-assessment, trial-level aggregation) focusing on the patient-centric task (i.e. not comparable to our trial-centric approach). The second stage, that is has in common with the trial-centric approach, leverages GPT-4 and achieves a 87\% criterion-level accuracy on the SIGIR 2016 cohort (\cite{koopman2016test}).

To evaluate the benefit of providing a rationale alongside the actual eligibility decision, \cite{wornow2024zeroshotclinicaltrialpatient} assessed the interpretability of LLM-generated assessments. Clinicians reviewed the rationales generated by the LLM for each eligibility decision, demonstrating that the model can produce coherent explanations for 97\% of its correct decisions and even 75\% of its incorrect ones.

\subsection{Shortcomings}
\label{sec:shortcomings}

Despite promising feasibility, above approaches fell short when applied in a real-world setting. This is mainly due to four challenges:
\begin{enumerate}
\item \textbf{Lack of basic reasoning skills, limiting assessment quality}: Despite the improvements made to chat models (a.k.a. single-turn models, i.e. models optimized for fast conversational interactions with humans as opposed to reasoning models optimized for problem solving), even the most advanced ones, like GPT-4o, fail at some simple tasks. For example, date-related calculations or assessing some logical expressions remains a challenge for them (e.g. Figure \ref{fig:wrong-assessment-date} shows GPT-4o struggling with simple date calculations or making a wrong decision on lab values check).
\begin{figure}[htp]
    \centering
    \includegraphics[width=\textwidth]{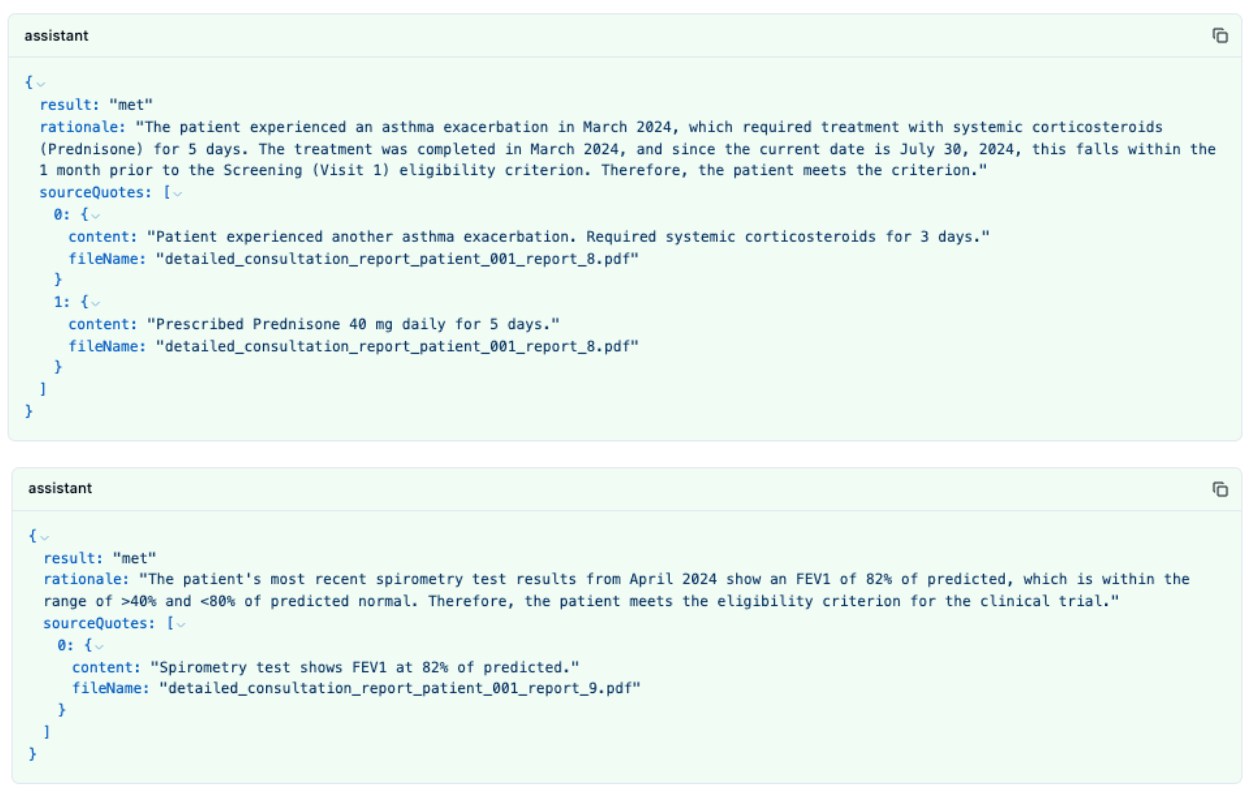}
    \caption{Examples of GPT-4o struggling with basic evaluation}
    \label{fig:wrong-assessment-date}
\end{figure}
\item \textbf{Reliance on expert-refined criteria, limiting broad applicability}: Many of the aforementioned approaches (\cite{wornow2024zeroshotclinicaltrialpatient}, \cite{wong2023scalingclinicaltrialmatching}) rely on having experts ask the right questions or refine the eligibility criteria to make them less ambiguous. While it is interesting in the context of one particular trial, it is not compatible with a broadly applicable pre-screening tool. Having an LLM autonomously refine a criteria could also be an option, but as eligibility criteria typically undergo rigorous scrutiny to meet many constraints, modifying them could lead to unintended consequences, so we put aside this approach.
\item \textbf{Lack of visual understanding, limiting medical record coverage}: Furthermore, despite the rise of EHRs, numerous institutions still rely heavily on handwritten notes. \cite{doi:10.1308/rcsann.2022.0066} showed that in a sample of 405 patient notes taken in 2020 and 2021 in the UK, 40\% of notes were handwritten. Besides, a lot of data, even if digitized, remains better represented in a visual form, like tables, charts and graphs. We took a sample of 1000 pages from our sites’ medical records and classified them using GPT-4o. We found that 46\% of them contained visually challenging items (see Figure \ref{fig:visualchallenges} for details of about the type of challenges. Prompt, and full results can be found in the Appendix \ref{sec:visual_exploration_appendix}). As a result, passing all this data into a text-only LLM does not allow to fully leverage the available information. Indeed, despite recent improvements, OCR and captioning pipelines are known to be expensive, time-consuming and error-prone. \cite{faysse2025colpaliefficientdocumentretrieval} showed that parsing + captioning of tables and graphs + embedding took more than 7 seconds per page while embedding the image of the page directly took less than half a second, while achieving better retrieval performance.
\begin{figure}[htp]
    \centering
    \includegraphics[width=\textwidth]{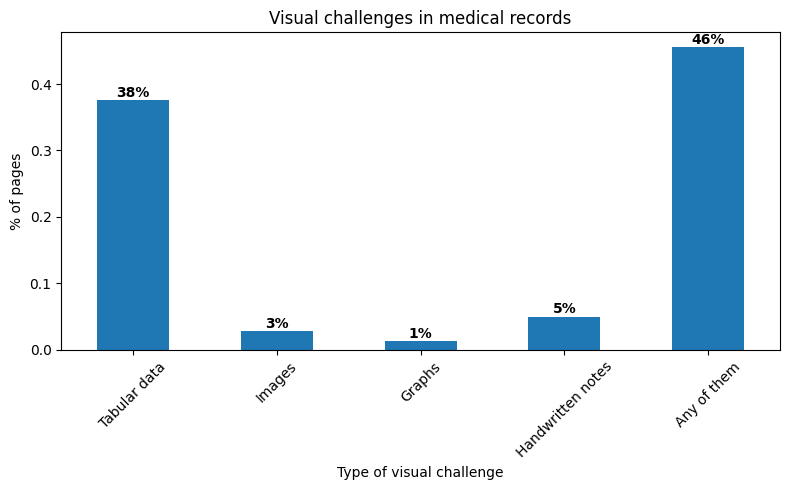}
    \caption{Visually challenging items found in medical records. \textit{Example: 38\% of medical records contain at least one table. 46\% of them contain at least one of the 4 types of challenges.}}
    \label{fig:visualchallenges}
\end{figure}
\item \textbf{Lack of a generic EHR integration, preventing universal deployments}: There is no generic way to interface with research centers and hospitals’ information systems. Within the US alone, the EHR vendor market is highly fragmented. \cite{SORACE2020104037} showed that only 4.5\% of Medicare beneficiaries had their meaningful-use (MU) medical records associated with a single vendor, and 19.8\% of them had their MU medical records stored in 8 or more vendors. And the problem is of course even more present at a global scale, e.g. in Europe where the top 15 EHRs accounted for only 58\% of the market in 2022 (\cite{Signify}). The lack of such a generic integration is an obstacle to the extraction of structured data or even textual data in general.
\end{enumerate}

\section{Methods}

\subsection{Ethical considerations}
This study was conducted using anonymized patient data that was de-identified in accordance with the HIPAA Safe Harbor method. De-identification was performed using Google Cloud Data Loss Prevention (GCP DLP), an automated tool that detects and removes Protected Health Information (PHI) from unstructured data. The de-identification process ensured the removal or transformation of 18 HIPAA-defined identifiers, including patient names, dates, addresses, and other uniquely identifiable information.

Because the dataset does not contain identifiable private information, this study was deemed exempt from IRB review in accordance with institutional policies and federal regulations. No individual-level re-identification was attempted, and all analyses were performed on the anonymized dataset. The data was handled in compliance with applicable ethical guidelines and regulatory standards to ensure patient privacy and confidentiality.

\subsection{Pipeline description}

Our study builds on previous research by introducing key elements that make the process more applicable to the real world.

In particular, we leverage three recent innovations to address the aforementioned limitations of previous attempts.

\begin{enumerate}

\item \textbf{Vision-Language models (VLMs)}: With the rise of large VLMs, it has become possible to implement a fully visual pipeline. Meaning that advanced models can now directly ingest images of medical records to make eligibility assessments, hence allowing to fully leverage the information available in medical records and bypassing the need for heavy and error-prone OCR pipelines.

\item \textbf{Visual retrieval}: Because of improvements made to the VLMs scene, extensive work has also been done to make retrieval tasks work in a fully visual manner. To address this need \cite{faysse2025colpaliefficientdocumentretrieval} introduced ColPali, a visual retriever, and while doing so they introduced ViDoRe, a benchmark to compare performances of visual retrieval methods. Then, \cite{VoyageAI} released a multimodal embedding model, relying on a single vector per image, making it much easier to scale, and outperforming all other models on the ViDoRe benchmark, making it a natural choice for us.

\item \textbf{Reasoning model}: Finally, the rise of the new reasoning model paradigm, introduced by OpenAI’s o1, made it possible to evaluate more complex criteria and have the model succeed at assessing complex logical expression or arithmetic tasks that chat models would fail at.
\end{enumerate}

\begin{figure}[htp]
    \centering
    \includegraphics[width=\textwidth]{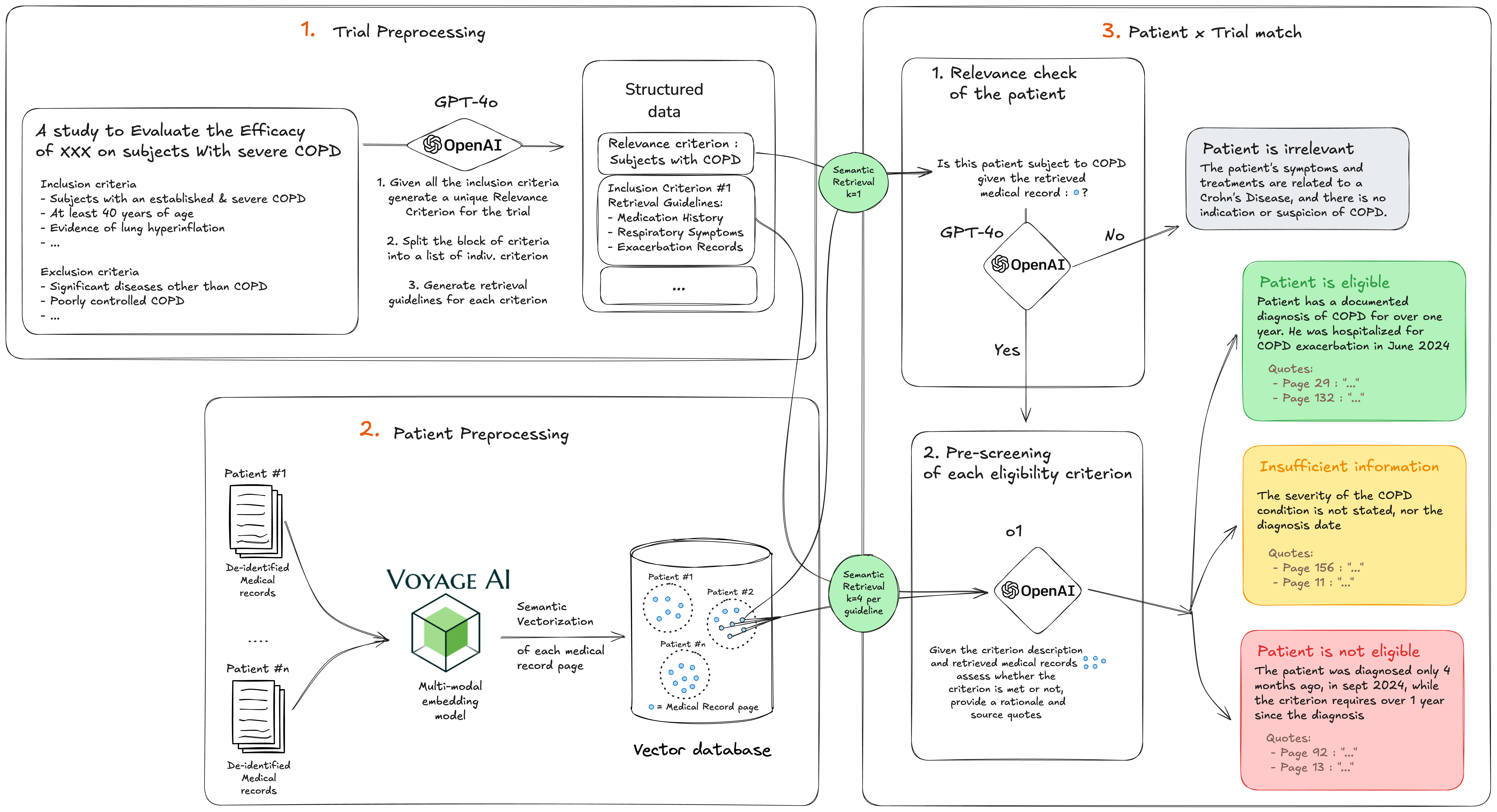}
    \caption{Our patient-trial matching pipeline}
    \label{fig:pipeline}
\end{figure}

Figure \ref{fig:pipeline} gives an overview of our whole pipeline, composed of three components: a trial preprocessing, a patient preprocessing, and a 2-step patient-trial matching. We describe below, each component in details.

\subsubsection{Trial preprocessing}
\label{sec:trialpreprocessing}
The goal of this step is to convert a block of free text eligibility criteria into a set of ready-to-assess and well structured criteria.
It is made of three independent LLM calls. All prompts can be found in Appendix \ref{sec:trial_preprocessing_appendix}.
\begin{enumerate}
    \item \textbf{Split criteria}: When retrieving the eligibility criteria (i.e. from clinicaltrials.gov), we get a free-text block. In order to evaluate these criteria one-by-one we split this block into individual criterion.
    \item \textbf{Generate relevance criterion}: Clinical sites databases usually contain many patients. For any given trial, most of them won't be relevant. E.g. a dermatology patient is irrelevant for a cardiovascular trial, so there is no need to pre-screen them for it, and we want to filter them upfront. For this purpose, we have an LLM generate a "relevance criterion", typically containing the main condition targeted by the trial. It will be used later in the process, to perform a low-cost relevance check for each trial x patient pair.
    \item \textbf{Generate retrieval guidelines}: We observed that, when confronted to large medical records, Clinical Researchers tend to make assumptions as to where they could find the relevant information. This is also recommended in \cite{sarkar2014conducting}. In this fashion, we have an LLM generate such retrieval guidelines.
\end{enumerate}

\subsubsection{Patient preprocessing}
We process uploaded files in the following way:
\begin{enumerate}
    \item \textbf{Split and de-identification}: We first split the PDF into individual images and anonymize them using Google Cloud DLP de-identification service to redact any patient-identifying information.
    \item \textbf{Embedding}: We send base64 encodings of these images to the multi-modal embedding model introduced by \cite{VoyageAI}.
    \item \textbf{Vector storing}: We store the resulting dense vectors into a vector database, for later semantic retrieval.
\end{enumerate}

\subsubsection{Patient x Trial matching}
The patient-trial match is composed of two steps. Prompts can be found in Appendix \ref{sec:patient_trial_matching_appendix}.
\begin{enumerate}
    \item \textbf{Relevance check}: In order to not waste resources on patients that are evidently not relevant to a given trial, we first make a low-cost relevance check. Using the relevance criterion generated at section \ref{sec:trialpreprocessing}, we make a semantic retrieval of the most relevant page of the patient's medical record, i.e. we embed the relevance criterion with the same embedding model and execute a similarity search on our vector DB. We submit that page to GPT-4o to evaluate whether the patient meets the relevance criterion.
    \item \textbf{Assessment}: If the patient passed the relevance check, it is then assessed against each eligibility criterion of the trial. To do so, we use the guidelines generated at section \ref{sec:trialpreprocessing} to semantically retrieve from the vector database the right parts of their medical records and submit them to a multi-modal reasoning model (OpenAI's o1) to obtain an assessment (Eligible / Not Eligible / Insufficient information).
\end{enumerate}

Throughout each phase of our pipeline, we opted for OpenAI’s latest models due to their consistent top performance across multiple benchmarks. However, we did not conduct a comprehensive benchmarking evaluation of LLM providers.

\subsection{User feedback}
\label{sec:feedback}

After the AI assessment is complete, users can confirm or rectify each criterion-level assessment and classify patients as "To screen" or "Not eligible", hence giving us valuable feedback to measure the quality of our assessments.

Figure \ref{fig:prescreening} below shows what the assessments look like in our pre-screening tool, and how users can give feedback.

\begin{figure}[htp]
    \centering
    \includegraphics[width=\textwidth]{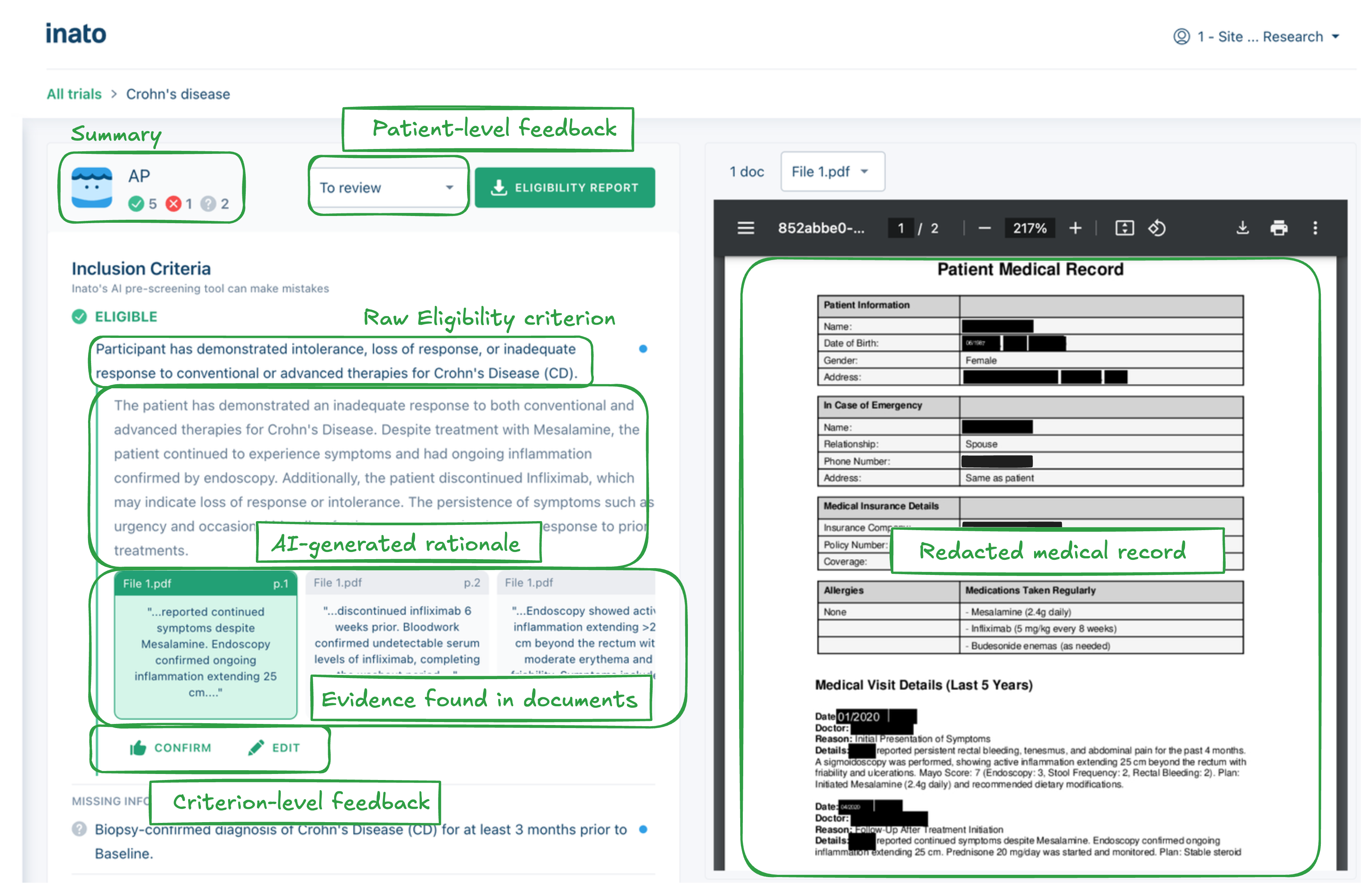}
    \caption{Screenshot of our prescreening tool - Patient pre-screened with rationale, source quotes and possibility to give feedback}
    \label{fig:prescreening}
\end{figure}

\section{Datasets}

We used two different datasets to evaluate our pipeline:

\begin{enumerate}
    \item \textbf{n2c2 public dataset}: We used the dataset from the Track 1 of the 2018 n2c2 cohort selection benchmark, which is the largest publicly available clinical trial patient matching benchmark (\cite{stubbs2019cohort}). It was also used in \cite{wornow2024zeroshotclinicaltrialpatient}, so it makes it a natural choice to compare our performances. Though, as it's a text-only dataset, we converted the text to low-resolution images to test the visual aspect of our pipeline .
    \item \textbf{our own dataset based on in-app user feeback}: Following the release of our tool, we had it beta tested by 30 sites. We used their in-app feedbacks to evaluate our performance. The methodology to build this dataset is described below.
\end{enumerate}

\subsection{n2c2 public dataset}

The 2018 n2c2 Clinical Trial Cohort Selection challenge dataset is composed of 288 unique diabetic patients, partitioned into a training set of 202 individuals and a test set of 86. Each patient has 2 to 5 clinical notes. These patients were originally drawn from a combined database at Massachusetts General Hospital and Brigham and Women’s Hospital. Thirteen commonly employed inclusion criteria were used (\cite{stubbs2019cohort}). These criteria can be found in Appendix \ref{sec:n2c2_criteria_appendix}. All patient data are provided in an unstructured format. To obtain the final labels, two medical experts independently annotated each patient’s status (whether the 13 criteria were met or not).

Due to the visual nature of our pipeline, we processed all clinical notes to convert them from text to image format. We chose a low-resolution of 72 dots-per-inch to be conservative on the quality of images we would receive in real life.

\subsection{Real world dataset}
\label{sec:rw_dataset}
As explained in section \ref{sec:feedback}, we gather feedback from our users in two different ways: 
\begin{enumerate}
    \item via direct confirmation/rectification of each criterion-level assessment
    \item at the patient level as they classify them as "To be screened" or "Not eligible"
\end{enumerate}

To build a gold-standard dataset for performance assessment, we used criterion-level reviews as ground truth whenever they were available. When only patient-level reviews were available, we relied on heuristics to deduce the true value of some criterion-level assessments. These heuristics are detailed in Appendix \ref{sec:heuristic}.

Using this method, we obtained \textbf{7021} labeled patient-criterion pairs, distributed across 30 sites, 36 distinct trials, 485 patients and 650 eligibility criteria. We further describe them in following subsections.

\subsubsection{Sites}
All 30 sites that beta tested our tool were located in the United States.

The single most active site represented 28\% of the labeled criterion-level assessments, and the five most active represented 86\% of them.

To qualify sites, we categorized them in the 6 categories reported in table \ref{tab:research_site_types}. Associated definitions can be found in Appendix \ref{sec:research_site_type_defintions_appendix}.

\begin{table}[h]
    \centering
    \begin{tabular}{lc}
        \hline
        \textbf{Type} & \textbf{Count} \\ \hline
        Independent research site embedded in an outpatient clinic & 11 \\ \hline
        Services company & 6 \\ \hline
        Research center part of a site network & 6 \\ \hline
        Independent research site & 5 \\ \hline
        Outpatient clinic & 1 \\ \hline
        Oncology care center & 1 \\ \hline
    \end{tabular}
    \caption{Summary of Research Site Types}
    \label{tab:research_site_types}
\end{table}

\subsubsection{Evaluators}
At each site, one person was in charge of reviewing the AI-generated assessments. We report in table \ref{tab:evaluators_primary_focus} their primary focus as they filled it in in their user profile.

\begin{table}[h]
    \centering
    \begin{tabular}{lc}
        \hline
        \textbf{Primary Focus} & \textbf{Count} \\ \hline
        Site Director & 12 \\ \hline
        Study Coordinator & 11 \\ \hline
        Other & 3 \\ \hline
        Recruitment Specialist & 3 \\ \hline
        Principal Investigator & 1 \\ \hline
    \end{tabular}
    \caption{Summary of evaluators's primary focus}
    \label{tab:evaluators_primary_focus}
\end{table}

\subsubsection{Trials}
The labels covered \textbf{36} different studies, evenly split between phase II and phase III trials. The focus of these trials was diverse, as shown in table \ref{tab:therapeutic_areas}. The top-3 therapeutic areas (Gastroenterology, Pulmonology and Dermatology) accounted for 77\% of all reviewed eligibility criteria.

\begin{table}[htbp]
\centering
\begin{tabular}{lrr}
\hline
\textbf{Therapeutic Area} & \textbf{Trials} & \textbf{Labeled eligibility criteria} \\
\hline
Pulmonology & 11 & 1305 \\
Gastroenterology & 4 & 2950 \\
Dermatology & 3 & 1162 \\
Cardiovascular & 3 & 590 \\
Endocrinology & 3 & 306 \\
Rheumatology & 2 & 51 \\
Psychiatry & 2 & 80 \\
Oncology & 2 & 7 \\
Hematology & 2 & 7 \\
Obesity & 1 & 101 \\
Pediatrics & 1 & 51 \\
Nephrology & 1 & 420 \\
Gynecology & 1 & 9 \\
\hline
\end{tabular}
\caption{Therapeutic areas }
\label{tab:therapeutic_areas}
\end{table}

\subsubsection{Eligibility criteria}
All eligibility criteria of these 36 trials were assessed by our tool, meaning that there was no hand-picking of the easy-to-assess criteria, resulting in a total of \textbf{650} criteria. To better qualify these criteria, we propose a 3-axis classification:
\begin{enumerate}
    \item \textbf{Domain}: Describes the general nature of the requirement (e.g., demographic detail, disease-specific factor, lab value, etc.).
    \item \textbf{Data Format}: Differentiates between criteria that can be captured as structured, numeric, or coded fields and those that rely on complex free-text.
    \item \textbf{Temporal Constraint}: Highlights whether the requirement applies within a particular time frame (e.g., “within 6 months” or “for at least 1 year”).
\end{enumerate}

The classification of every criterion was done by GPT-4o. We present below the results of this classification.

\paragraph{Domain}
We assign each criterion to its most relevant functional “domain”. These domains capture what the criterion is about. The prompt containing the definition of each domain and used to make this classification can be found in Appendix \ref{sec:domain_classification_appendix}. Table \ref{tab:domain_stats} gives the distribution of eligibility criteria per domain.

\begin{table}[h]
    \centering
    \begin{tabular}{lrr}
    \textbf{Domain} & \textbf{Exclusion crit.} & \textbf{Inclusion crit.} \\ \hline
    Comorbidity or Medical History & 1224 (183) & 248 (15) \\ \hline
    Demographic / Administrative  & 23 (2)  & 303 (32) \\ \hline
    Disease or Condition Specific  & 126 (16) & 587 (40) \\ \hline
    Lab or Biomarker              & 322 (35) & 269 (25) \\ \hline
    Other Pragmatic              & 29 (5)  & 213 (15) \\ \hline
    Performance or Functional Status & 7 (4) & 33 (10) \\ \hline
    Prior or Concomitant Treatments & 1326 (142) & 277 (24) \\ \hline
    Safety or Risk                & 1086 (96) & 6 (5)  \\ \hline
    \end{tabular}
    \caption{Number of criteria assessments per domain (number of unique criteria)}
    \label{tab:domain_stats}
\end{table}

\paragraph{Requested Data format}
We further classify criteria by the type of data they require to be assessed. The first group consists of criteria that can be evaluated using discrete or numeric data fields, codes, or straightforward yes/no inputs (we call these criteria \textbf{Structured}). And the other group contains all other criteria, i.e. those that require the interpretation of free-text medical notes or narratives (we call these criteria \textbf{Unstructured}). The prompt used to make this classification can be found in Appendix \ref{sec:data_format_classification_appendix}. Table \ref{tab:data_format_stats} gives the distribution of eligibility criteria per requested data format.

\begin{table}[h]
    \centering
    \begin{tabular}{lrr}
    \textbf{Domain} & \textbf{Exclusion crit.} & \textbf{Inclusion crit.} \\ \hline
    Structured & 967 (105) & 981 (89) \\ \hline
    Unstructured  & 3185 (379)  & 955 (77) \\ \hline
    \end{tabular}
    \caption{Number of criteria assessments per requested data format (number of unique criteria)}
    \label{tab:data_format_stats}
\end{table}

\paragraph{Temporal constraint}
Finally, we classify criteria depending on whether they mention a particular time frame (e.g., “within 6 months” or “for at least 1 year”). The prompt used to make this classification can be found in Appendix \ref{sec:temporal_classification_appendix}. Table \ref{tab:temporal_stats} gives the distribution of eligibility criteria per presence of temporal constraint.

\begin{table}[h]
    \centering
    \begin{tabular}{lrr}
    \textbf{Domain} & \textbf{Exclusion crit.} & \textbf{Inclusion crit.} \\ \hline
    Yes & 1905 (211) & 660 (57) \\ \hline
    No  & 2247 (273)  & 1276 (109) \\ \hline
    \end{tabular}
    \caption{Number of criteria assessments per temporal constraint (number of unique criteria)}
    \label{tab:temporal_stats}
\end{table}

\subsubsection{Medical records}
In total, the medical records of \textbf{485} distinct patients were uploaded into our tool.

Each patient was on average associated with 1.6 medical records (i.e. PDF files): most patients (69\% of them) had only one medical record, but some of them had up to 9 medical records associated.

On average, medical records were \textbf{28} pages long. 50\% of them were 12 pages or more and 10\% of them were 73 pages or more.

To better qualify the content of medical records, we used the following categorization:
\begin{enumerate}
    \item \textbf{Administrative Documents:} These include forms and records capturing patient identification, insurance details, and legal consent (e.g., registration forms and HIPAA notices).
    \item \textbf{Clinical Notes:} This category encompasses encounter documentation such as History and Physical reports, progress notes, consultation notes, and discharge summaries, along with transition of care documents.
    \item \textbf{Diagnostic Reports:} These consist of laboratory, imaging, and pathology reports that present clinical test results and diagnostic findings (e.g., CBCs, CT scans, biopsy reports).
    \item \textbf{Procedural \& Treatment Documents:} This group includes procedure reports (e.g., operative, endoscopy, and interventional reports) as well as medication orders and treatment plans.
\end{enumerate}

To understand the distribution of our dataset, we randomly sampled 100 medical records, broke them down into 2947 individual images and used an LLM to classify each page into one of these 4 categories. The result of this categorization is shown in table \ref{tab:medical_records_distribution} below.

\begin{table}[h]
    \centering
    \begin{tabular}{lrr}
        \hline
        \textbf{Type} & \textbf{Pages} & \textbf{Medical records where present} \\ \hline
        Clinical Notes & 1519 & 85 \\ \hline
        Diagnostic Reports & 785 & 69 \\ \hline
        Procedural and Treatment Documents & 467 & 69 \\ \hline
        Administrative Documents & 176 & 51 \\ \hline
        Total & 2947 &  \\ \hline
    \end{tabular}
    \caption{Distribution of Medical Records types. \textit{Example: Out of 2947 pages analyzed, 785 of them were diagnostic reports. This type of content was present in 69 of the 100 medical records analyzed.}}
    \label{tab:medical_records_distribution}
\end{table}

\section{Results}

\subsection{Results on the n2c2 dataset}

Table \ref{tab:classification_report_overall_n2c2} below showcases the results of our pipeline on the n2c2 dataset.

\begin{table}[!h]
\centering
\begin{tabular}{lccccc}
\hline
Class & Precision & Recall & F1-score & Sample Size \\
\hline
met & 0.92 & 0.93 & 0.93 & 1041 \\
unmet & 0.94 & 0.93 & 0.93 & 1325 \\
\hline
Accuracy &&&0.93&2366\\
Macro avg & 0.93 & 0.93 & 0.93 & 2366 \\
Weighted avg & 0.93 & 0.93 & 0.93 & 2366 \\
\hline
\end{tabular}
\caption{Criterion-level classification report of our visual o1-based pipeline on the n2c2 dataset converted in low-resolution images, leveraging all available clinical notes.}
\label{tab:classification_report_overall_n2c2}
\end{table}

\subsection{Results on our real-world dataset}

Table \ref{tab:classification_report_overall_rw} below showcases the results of our pipeline on real world data, as detailed in section \ref{sec:rw_dataset} above.

\begin{table}[htbp]
\centering
\begin{tabular}{lccccc}
\hline
Class & Precision & Recall & F1-score & Sample Size \\
\hline
met & 0.72 & 0.76 & 0.74 & 1683 \\
unknown & 0.97 & 0.99 & 0.98 & 1999 \\
unmet & 0.88 & 0.85 & 0.86 & 3339 \\
\hline
Accuracy & & & 0.87 & 7021\\
Macro avg & 0.86 & 0.86 & 0.86 & 7021 \\
Weighted avg & 0.86 & 0.87 & 0.87 & 7021 \\
\hline
\end{tabular}
\caption{Criterion-level classification report of our visual o1-based pipeline on our real-world dataset, leveraging the top-3 clinical notes, with guidelines activated.}
\label{tab:classification_report_overall_rw}
\end{table}

\subsubsection{Inference efficiency metrics}

When assessing criteria using all clinical notes, our pipeline took on average \(24.8 s\) per criterion, with an interquartile range of \(18.5s - 28.2s\).
This average fell to \(19.0 s\) per criterion when using only the top-3 clinical notes for each retrieval guideline.

As for the inference cost, our pipeline yielded a criterion-assessment for an average cost of \(\$0.15\). That cost decreased to \(\$0.09\) when using only the top-3 clinical notes for each retrieval guideline.

\subsubsection{User review efficiency metric}
\label{sec:user_efficiency}

As our paper focuses on enabling real-world productivity gains, we measured the time users took to review a complete patient--trial pair.

The median time observed to review a patient was \textbf{5.5 min}, the mean was around \textbf{9 min}, and the interquartile range spanned from 3 to 11 minutes. Heuristics used to obtain these metrics are described in Appendix \ref{sec:time_heuristics_appendix}.

\section{Discussion}

\subsection{Interpretation}

\subsubsection{Comparison of results}

By comparing our results shown in Table \ref{tab:classification_report_overall_n2c2} with previous attempts made with GPT-4 (Precision \textbf{0.91}, Recall \textbf{0.92}, \cite{wornow2024zeroshotclinicaltrialpatient}), we demonstrate that our o1 visual pipeline outperforms the previous SOTA despite relying on a degraded version of the dataset (resulting from the conversion of text clinical notes into low-resolution images). This advantage becomes especially clear when excluding previous approaches that relied on refined criteria (in which case GPT-4 achieves a precision of 0.89 and a recall of 0.74, \cite{wornow2024zeroshotclinicaltrialpatient}) — an aspect we consider non-scalable, as explained in section \ref{sec:shortcomings}.

Table \ref{tab:classification_report_overall_rw} presents the overall performance achieved by our pipeline on our real world dataset, showcasing an overall accuracy of 87\%, slightly below what we witnessed on the n2c2 public dataset. In section \ref{sec:lower_accuracy_reason} below, we give a potential explanation of this phenomenon.

In section \ref{sec:user_efficiency}, we show that the time required for prescreening using our tool is over 80\% more efficient than traditional manual reviews, as noted in \cite{penberthy2012effort}. However, this comparison is limited to the prescreening phase only, and does not reflect the complete screening process which also includes patient interactions.

\subsubsection{Further analysis}

We further analyze three aspects of our results:
\begin{enumerate}
    \item Reason for lower accuracy on real-world results
    \item Effect of retrieval and guidelines
    \item Accuracy grouped by criterion domain
\end{enumerate}

\paragraph{Reason for lower accuracy observed on real-world results}
\label{sec:lower_accuracy_reason}
To shed some more light and understand why the accuracy observed in our real world experiment is lower than the one obtained on the public dataset, we present separately the classification reports of inclusion (Table \ref{tab:classification_report_inclusion}) and exclusion criteria (Table \ref{tab:classification_report_exclusion}). This way, we observe that the majority of mislabeled samples (\textgreater 80\%) are coming from patients that we classified as Ineligible either due to an unmet inclusion criterion or a met exclusion criterion. This, along with user feedback analysis, possibly shows that CRCs using the tool are reluctant to exclude a patient without doing further investigations.
Both the met inclusion criteria and unmet exclusion criteria categories achieved a very high precision of 1.00.

\begin{table}[htbp]
\centering
\begin{tabular}{lcccc}
\hline
Class & Precision & Recall & F1-score & Sample Size \\
\hline
met     & 1.00 & 0.88 & 0.93 & 798 \\
unknown & 0.94 & 0.94 & 0.96 & 518 \\
unmet   & 0.86 & 0.97 & 0.91 & 562 \\
\hline
Accuracy     &      &      & 0.93 & 1878 \\
Macro avg    & 0.93 & 0.94 & 0.93 & 1878 \\
Weighted avg & 0.94 & 0.93 & 0.93 & 1878 \\
\hline
\end{tabular}
\caption{Subset classification report on inclusion criteria only - Real World dataset}
\label{tab:classification_report_inclusion}
\end{table}

\begin{table}[htbp]
\centering
\begin{tabular}{lcccc}
\hline
Class & Precision & Recall & F1-score & Sample Size \\
\hline
met     & 0.59 & 0.99 & 0.74 & 411 \\
unknown & 0.63 & 0.99 & 0.77 & 1327 \\
unmet   & 1.00 & 0.88 & 0.93 & 2360 \\
\hline
Accuracy     &      &      & 0.86 & 4098 \\
Macro avg    & 0.86 & 0.95 & 0.81 & 4098 \\
Weighted avg & 0.95 & 0.86 & 0.90 & 4098 \\
\hline
\end{tabular}
\caption{Subset classification report on exclusion criteria only - Real World dataset}
\label{tab:classification_report_exclusion}
\end{table}

\paragraph{Effect of retrieval and guidelines}

To measure the effect of retrieval and guidelines, we first measure the effect of having a semantic retrieval steps rather than using all clinical notes, on the n2c2 dataset. Additionally, we measure the performance of guidelines vs no guidelines.

\begin{figure}[htbp]
    \centering
    \begin{subfigure}{0.45\textwidth}
        \centering
        \includegraphics[width=\linewidth]{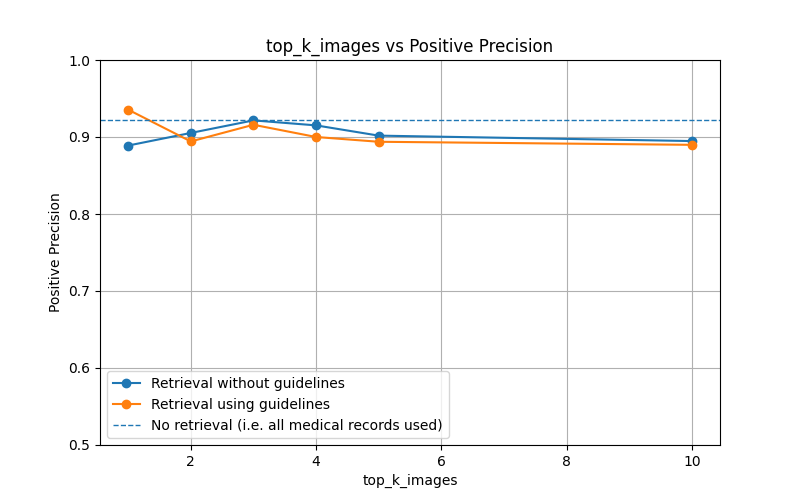}
        \caption{Precision}
        \label{fig:sub1}
    \end{subfigure}
    \hfill
    \begin{subfigure}{0.45\textwidth}
        \centering
        \includegraphics[width=\linewidth]{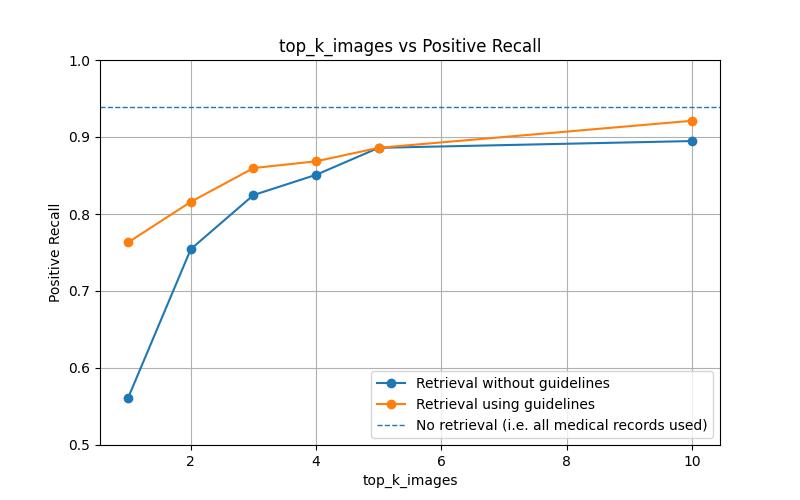}
        \caption{Recall}
        \label{fig:sub2}
    \end{subfigure}
    \caption{Precision and recall per retrieval strategy on n2c2 dataset assessments}
    \label{fig:metrics_per_retrieval_strategies}
\end{figure}

Figure \ref{fig:metrics_per_retrieval_strategies} shows that recall improves as we use more images. But interestingly, the precision is more or less always the same, even for a low number of images retrieved. This is likely explained by the fact that the n2c2 dataset is composed exclusively of inclusion criteria, and the LLM never considers that criteria are met if there is no tangible proof, resulting in a low number of false positives.
Besides, we notice that none of the retrieval strategies are good enough to match the performance of the all-notes run. This was already observed by \cite{wornow2024zeroshotclinicaltrialpatient}.

\begin{figure}[htbp]
    \centering
    \includegraphics[width=\textwidth]{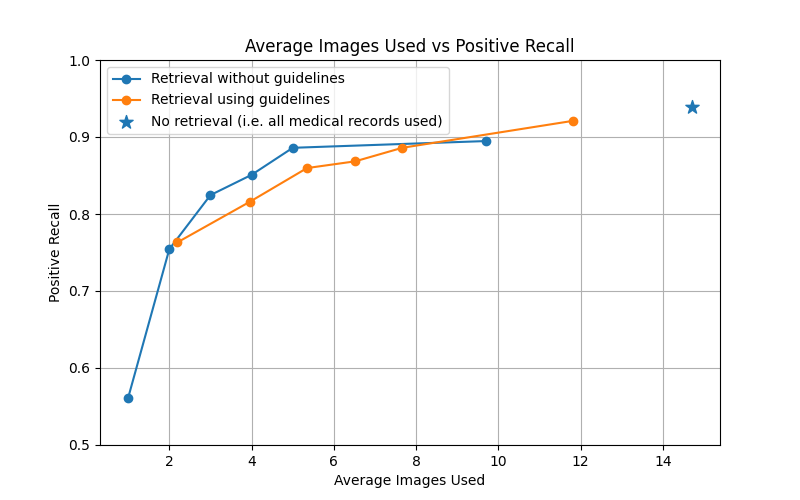}
    \caption{Recall per average images used for each retrieval strategy}
    \label{fig:recall_avg_images_used}
\end{figure}

The use of guidelines seems beneficial on the recall, but the comparison is unfair because when using guidelines, K images are retrieved for \textbf{each} guideline, and there are several guidelines per criterion, so on average more images are used for each criterion assessment. To visualize the real effect of guidelines, we plot the recall according to the average number of images used by each retrieval strategy. Figure \ref{fig:recall_avg_images_used} shows that the benefit is actually almost inexistent or even detrimental. We were surprised by this result as having research guideline is a clearly beneficial technique for manual patient review. Because we did not have a labeled retrieval dataset, we could not investigate further, but we believe there is still significant room for improvement in this area—particularly in the use of dynamic retrieval, where the model could successively retrieve pages as it reasons, rather than performing a single upfront retrieval step.

\subsubsection{Accuracy by criterion domain}

In Table \ref{tab:domain_accuracy} below we share the accuracy for each domain, as described in Appendix \ref{sec:domain_classification_appendix}. There is no large variability between domains.

\begin{table}[htbp]
\centering
\begin{tabular}{lrr}
\hline
\textbf{Domain} & \textbf{Accuracy} & \textbf{Samples} \\
\hline
Comorbidity or Medical History & 0.86 & 1792 \\
Demographic / Administrative & 0.89 & 352 \\
Disease or Condition Specific & 0.88 & 886 \\
Lab or Biomarker & 0.87 & 747 \\
Other Pragmatic & 0.97 & 229 \\
Performance or Functional Status & 0.84 & 32 \\
Prior or Concomitant Treatments & 0.86 & 1933 \\
Safety or Risk & 0.87 & 1112 \\
\hline
\end{tabular}
\caption{Accuracy across domains}
\label{tab:domain_accuracy}
\end{table}

\subsection{Limitations}

\subsubsection{Difficulty to calibrate firm vs uncertain assessments}
The analysis of user feedback indicates that CRCs tend to err on the side of caution: many are hesitant to classify a patient as definitively not eligible without conducting their own further verification. Consequently, defining a one-size-fits-all threshold for deciding when criteria are fully met or unmet rather than unknown remains an open question, particularly for incomplete medical records.

\subsubsection{Difficult to propose patient-level recommandations}
Similarly, aggregating multiple criterion-level decisions into a single, patient-level assessment remains nontrivial. We explored several approaches around absolute count and percentage of failing/succeeding criteria, but none of them offered an acceptable patient-level precision/recall trade-off.
Therefore, while the pipeline offers significant improvements at the criterion level, future efforts could further refine how these individual decisions merge into a final patient-level verdict.

\section{Code availability}
All the required code to replicate our results on the public n2c2 dataset is publicly available at \url{https://github.com/inato/ai_trial_eligibility_assessment} 

\newpage

\printbibliography

@misc{wong2023scalingclinicaltrialmatching,
      title={Scaling Clinical Trial Matching Using Large Language Models: A Case Study in Oncology}, 
      author={Cliff Wong and Sheng Zhang and Yu Gu and Christine Moung and Jacob Abel and Naoto Usuyama and Roshanthi Weerasinghe and Brian Piening and Tristan Naumann and Carlo Bifulco and Hoifung Poon},
      year={2023},
      eprint={2308.02180},
      archivePrefix={arXiv},
      primaryClass={cs.CL},
      url={https://arxiv.org/abs/2308.02180}, 
}

@misc{hamer2023improvingpatientprescreeningclinical,
      title={Improving Patient Pre-screening for Clinical Trials: Assisting Physicians with Large Language Models}, 
      author={Danny M. den Hamer and Perry Schoor and Tobias B. Polak and Daniel Kapitan},
      year={2023},
      eprint={2304.07396},
      archivePrefix={arXiv},
      primaryClass={cs.LG},
      url={https://arxiv.org/abs/2304.07396}, 
}

@misc{wornow2024zeroshotclinicaltrialpatient,
      title={Zero-Shot Clinical Trial Patient Matching with LLMs}, 
      author={Michael Wornow and Alejandro Lozano and Dev Dash and Jenelle Jindal and Kenneth W. Mahaffey and Nigam H. Shah},
      year={2024},
      eprint={2402.05125},
      archivePrefix={arXiv},
      primaryClass={cs.CL},
      url={https://arxiv.org/abs/2402.05125}, 
}

@article{Jin_2024,
   title={Matching patients to clinical trials with large language models},
   volume={15},
   ISSN={2041-1723},
   url={http://dx.doi.org/10.1038/s41467-024-53081-z},
   DOI={10.1038/s41467-024-53081-z},
   number={1},
   journal={Nature Communications},
   publisher={Springer Science and Business Media LLC},
   author={Jin, Qiao and Wang, Zifeng and Floudas, Charalampos S. and Chen, Fangyuan and Gong, Changlin and Bracken-Clarke, Dara and Xue, Elisabetta and Yang, Yifan and Sun, Jimeng and Lu, Zhiyong},
   year={2024},
   month=nov }

@article{SORACE2020104037,
    title = {Quantifying the competitiveness of the electronic health record market and its implications for interoperability},
    journal = {International Journal of Medical Informatics},
    volume = {136},
    pages = {104037},
    year = {2020},
    issn = {1386-5056},
    doi = {https://doi.org/10.1016/j.ijmedinf.2019.104037},
    url = {https://www.sciencedirect.com/science/article/pii/S1386505619310135},
    author = {James Sorace and Hui-Hsing Wong and Thomas DeLeire and Dashi Xu and Sheila Handler and Bruno Garcia and Thomas MaCurdy},
}

@article{doi:10.1308/rcsann.2022.0066,
    author = {Ekowo, O and Hammenga, C and Altaf, K and Chan, K and Bhardwaj, R and Nada, H},
    title = {A cross-sectional retrospective study comparing handwritten operation notes with electronic operation notes},
    journal = {The Annals of The Royal College of Surgeons of England},
    volume = {105},
    number = {1},
    pages = {35-42},
    year = {2023},
    doi = {10.1308/rcsann.2022.0066},
        note ={PMID: 35950972},
    
    URL = {
            https://doi.org/10.1308/rcsann.2022.0066
    },
    eprint = { 
            https://doi.org/10.1308/rcsann.2022.0066
    }
    ,
}

@misc{faysse2025colpaliefficientdocumentretrieval,
      title={ColPali: Efficient Document Retrieval with Vision Language Models}, 
      author={Manuel Faysse and Hugues Sibille and Tony Wu and Bilel Omrani and Gautier Viaud and Céline Hudelot and Pierre Colombo},
      year={2025},
      eprint={2407.01449},
      archivePrefix={arXiv},
      primaryClass={cs.IR},
      url={https://arxiv.org/abs/2407.01449}, 
}

@online{Signify,
  author    = {Alex Green, Signify Research},
  title     = {Dedalus and Oracle Cerner Occupy Top Spot for 2022 EHR Revenues in EMEA},
  year      = {2023},
  url       = {https://www.signifyresearch.net/insights/dedalus-oracle-cerner-occupy-top-spot-for-2022-ehr-revenues-in-emea/},
}

@online{VoyageAI,
  author    = {VoyageAI},
  title     = {voyage-multimodal-3: all-in-one embedding model for interleaved text, images, and screenshots},
  year      = {2024},
  url       = {https://blog.voyageai.com/2024/11/12/voyage-multimodal-3/},
}

@article{sarkar2014conducting,
  title={Conducting Record Review Studies in Clinical Practice},
  author={Sarkar, Siddharth and Seshadri, Divya},
  journal={Journal of Clinical and Diagnostic Research},
  volume={8},
  number={9},
  pages={JG01--JG04},
  year={2014},
  publisher={JCDR Research and Publications},
  doi={10.7860/JCDR/2014/8301.4806},
  url={https://www.ncbi.nlm.nih.gov/pmc/articles/PMC4225918/}
}

@article{stubbs2019cohort,
  title={Cohort selection for clinical trials: n2c2 2018 shared task track 1},
  author={Stubbs, Amber and Filannino, Michele and Soysal, Ergin and Henry, Samuel and Uzuner, Özlem},
  journal={Journal of the American Medical Informatics Association},
  volume={26},
  number={11},
  pages={1163--1171},
  year={2019},
  publisher={Oxford University Press},
  doi={10.1093/jamia/ocz163},
  url={https://pubmed.ncbi.nlm.nih.gov/31562516/}
}

@article{shah2021accelerating,
  title={Accelerating pre-formulation investigations in early drug product life cycles using predictive methodologies and computational algorithms},
  author={Shah, Harsh S and Chaturvedi, Kaushalendra and Kuang, Shanming and Wang, Jian},
  journal={Therapeutic Delivery},
  volume={12},
  number={11},
  pages={789--797},
  year={2021},
  publisher={Future Science},
  doi={10.4155/tde-2021-0043},
  url={https://pubmed.ncbi.nlm.nih.gov/34792419/}
}

@article{penberthy2012effort,
  title={Effort Required in Eligibility Screening for Clinical Trials},
  author={Penberthy, Lynne T. and Dahman, Bassam A. and Petkov, Valentina I. and DeShazo, Jonathan P.},
  journal={Journal of Oncology Practice},
  volume={8},
  number={6},
  pages={365--370},
  year={2012},
  publisher={American Society of Clinical Oncology},
  doi={10.1200/JOP.2012.000646},
  url={https://www.ncbi.nlm.nih.gov/pmc/articles/PMC3500483/}
}

@misc{koopman2016test,
  author       = {Koopman, Bevan and Zuccon, Guido},
  title        = {A Test Collection for Matching Patient to Clinical Trials},
  year         = {2016},
  version      = {v4},
  publisher    = {CSIRO},
  doi          = {10.4225/08/58e2e83d92c2b},
  url          = {https://doi.org/10.4225/08/58e2e83d92c2b}
}

\newpage

\appendix

\section{Exploration of medical records with visual elements}
\label{sec:visual_exploration_appendix}
Prompt used to process a sample of medical record images and assess whether they contain visual elements:
\begin{verbatim}
You are an AI expert, your job is to decide whether a given screenshot 
of document contains complex visual information that would require
the use of a vision-language model to fully leverage
the information available

Give your answer in JSON format with following keys:

- rationale : your rationale
- visual_elements: a list of visual elements present in the image. Each 
element should be one of the following values : "Tabular data", 
"Images", "Graphs", "Handwritten notes", "Other". Leave empty if 
no visual elements are present.

Additional instructions:
- A handwritten signature is not considered handwritten notes.
\end{verbatim}

\section{Trial preprocessing}
\label{sec:trial_preprocessing_appendix}
\subsection{Criteria split}
Prompt used to split a block of text describing all the criteria into a structured JSON object:
\begin{verbatim}
## Task
You will be provided with a list of eligibility criteria from 
a clinical trial.
Your job is to structure this data into JSON format.

## Additional instructions
- If a criterion given to you is nested, meaning that it contains 
several conditions, you must keep it in a single criterion in your answer.
You must explain the criterion in plain English.
\end{verbatim}

\subsection{Guidelines generation}
Prompt used to generate retrieval guidelines:
\begin{verbatim}
# Job
You are a senior Clinical Research Coordinator. 
Your job is to help a fellow CRC by telling him what he should 
be looking for in patient medical record to assess a given 
eligibility criterion.

# Task
You will be given an eligibility criterion, you must output 
between 1 and 4 guidelines to guide the retrieval of the right
pages of a medical records.
Your answers will be used by a RAG system to retrieve the right
pages from a medical record.

# Format
Format your response as a JSON object with the following keys:
* guidelines: Array of string - A set of guidelines to find 
the right pages of a medical record given the criterion.

\end{verbatim}

\subsection{Relevance criterion generation}
Prompt used to generate the criterion used to assess the relevance of patients for a given trial:
\begin{verbatim}
### Role
You are a knowledgeable and detail-oriented Clinical Researcher 
with deep expertise in clinical trials.

### Objective
Your job is to simplify the work of a fellow Clinical Researcher 
by helping them find relevant patients for a given clinical trial.

### Task
Based on a clinical trial title and its eligibility criteria, 
generate a single "diagnosis" criterion that will allow your 
colleague to filter a wide pool of patients in order to retain
only the relevant ones for the trial. Your criterion should 
be wide enough so that it encompasses all patient that could
be relevant for the trial. It should also be simple and only
target a single condition.

### Output Format
Respond in JSON format with the following structure:
- patientRelevanceCriterion : str

### Guidelines
- Ensure your criterion is lenient enough to not miss any 
relevant patient

### Trial
Here is the clinical trial you should consider:
{{trial_name}}

### Inclusion criteria
And here are the inclusion criteria of this trial:
{{inclusion_criteria}}

\end{verbatim}

\section{Patient trial matching}
\label{sec:patient_trial_matching_appendix}
Prompt used to assess whether a patient is eligible to a given criterion:
\begin{verbatim}
# Job
You are a senior Clinical Research Coordinator. Your job is to 
assess the eligibility of a patient to a given eligibility criterion
in the context of a clinical trial.

# Task
You will be given a set of medical records as well as an eligibility
criterion to assess. You must decide whether that patient meets the 
criterion.

# Format
Format your response as a JSON object with the following keys:
* rationale: str - Your reasoning as to why the patient does/does
not meet the criterion.
* is_met: bool - Whether the criterion is met or not

# Additional instructions:
* When the criterion refers to a specific period you must 
be particularly careful about the current date.

# Eligibility Criterion
The eligibility criterion being assessed is:
{criterion_description}

# Current Date
Assume that the current date is: {assessment_as_of_date}
\end{verbatim}

\section{n2c2 eligibility criteria}
\label{sec:n2c2_criteria_appendix}
Below are the eligibility criteria present in the n2c2 dataset, taken from \cite{stubbs2019cohort}:
\begin{enumerate}
    \item \textbf{DRUG-ABUSE}: Drug abuse, current or past
    \item \textbf{ALCOHOL-ABUSE}: Current alcohol use over weekly recommended limits
    \item \textbf{ENGLISH}: Patient must speak English
    \item \textbf{MAKES-DECISIONS}: Patient must make their own medical decisions
    \item \textbf{ABDOMINAL}: History of intra-abdominal surgery, small or large intestine resection, or small bowel obstruction
    \item \textbf{MAJOR-DIABETES}: Major diabetes-related complication. For the purposes of this annotation, we define “major complication” (as opposed to “minor complication”) as any of the following that are a result of (or strongly correlated with) uncontrolled diabetes:
    \begin{itemize}
        \item Amputation
        \item Kidney damage
        \item Skin conditions
        \item Retinopathy
        \item Nephropathy
        \item Neuropathy
    \end{itemize}
    \item \textbf{ADVANCED-CAD}: Advanced cardiovascular disease (CAD). For the purposes of this annotation, we define “advanced” as having 2 or more of the following:
    \begin{itemize}
        \item Taking 2 or more medications to treat CAD
        \item History of myocardial infarction (MI)
        \item Currently experiencing angina
        \item Ischemia, past or present
    \end{itemize}
    \item \textbf{MI-6MOS}: MI in the past 6 months
    \item \textbf{KETO-1YR}: Diagnosis of ketoacidosis in the past year
    \item \textbf{DIETSUPP-2MOS}: Taken a dietary supplement (excluding vitamin D) in the past 2 months
    \item \textbf{ASP-FOR-MI}: Use of aspirin to prevent MI
    \item \textbf{HBA1C}: Any hemoglobin A1c (HbA1c) value between 6.5\% and 9.5\%
    \item \textbf{CREATININE}: Serum creatinine $\ge$ upper limit of normal
\end{enumerate}

\section{Heuristic to deduce ground truth from user feedback}
\label{sec:heuristic}
Below is the decision tree we used to infer a ground truth dataset from the feedback collected in our app.

\begin{figure}[htp]
    \centering
    \includegraphics[width=\textwidth]{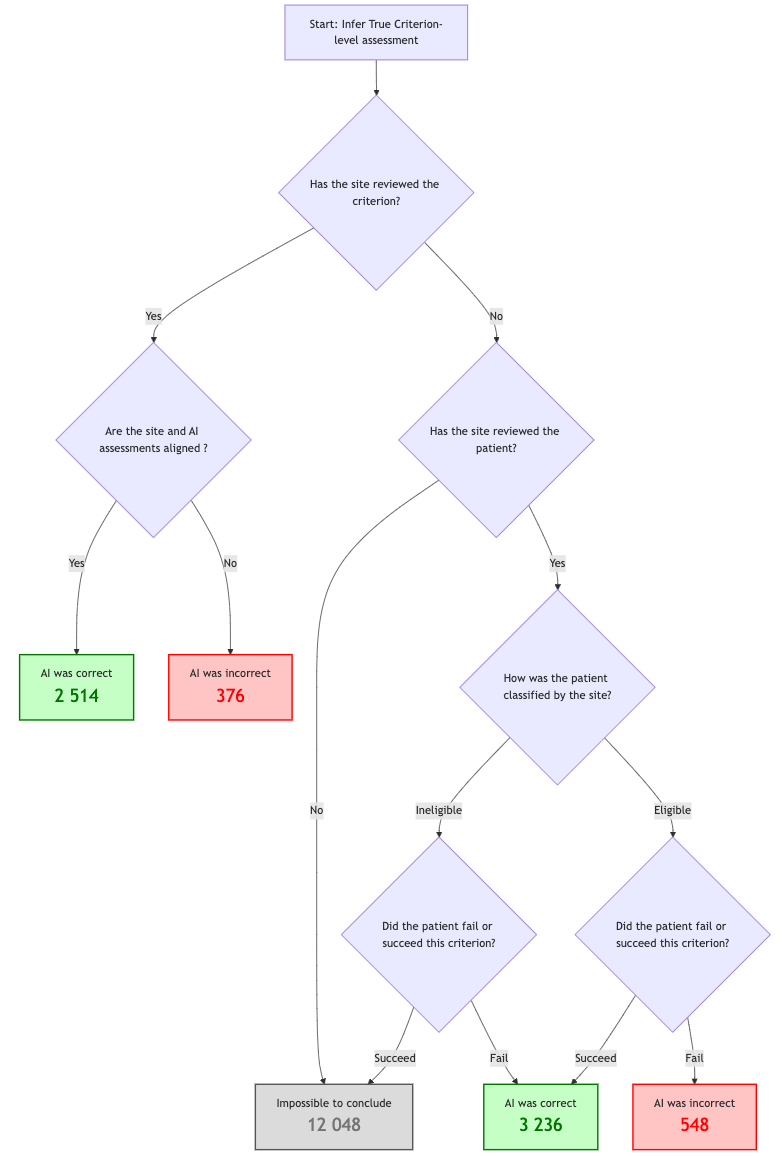}
    \caption{Decision tree to infer ground truth from user feedback}
    \label{fig:decision}
\end{figure}

\section{Classification of eligibility criteria: domain}
\label{sec:domain_classification_appendix}
Below is the prompt used to classify eligibility criteria by domain:
\begin{verbatim}
You are a domain-classification assistant for clinical trial eligibility criteria. 

**Task**: 
You will receive exactly one clinical trial eligibility criterion. 
Your job is to decide which one of the following domains best describes it:

1. **Demographic/Administrative** 
 - Typically involves age, sex, race, location, or administrative details 
 (e.g., "Must be >18 years old," "Female participants only," or "Must live 
 within 50 miles of the site").
2. **DiseaseOrConditionSpecific** 
 - Directly relates to the condition under study (e.g., "Histologically confirmed diagnosis
 of breast cancer," "Meets the 2010 ACR/EULAR criteria for RA").
3. **ComorbidityOrMedicalHistory** 
 - Pertains to other medical conditions or history (e.g., "History of diabetes," "Any prior 
 cardiac event," "Present mental health disorder").
4. **PriorOrConcomitantTreatments** 
 - Concerns the use of medications, treatments, surgeries, or therapies (e.g., "No prior 
 use of biologics," "Must be on stable dose of corticosteroids").
5. **LabOrBiomarker** 
 - Involves specific lab-value thresholds or biomarkers (e.g., "ALT or AST <3× ULN"
 , "Hemoglobin >10 g/dL").
6. **PerformanceOrFunctionalStatus** 
 - References a performance scale or functional capacity (e.g., ECOG performance status
 , Karnofsky scale, or the ability to perform daily activities).
7. **SafetyOrRisk** 
 - Addresses safety-related factors (e.g., active infection risk, pregnancy status, known
 severe allergies, or any condition that could endanger the patient).
8. **OtherPragmatic** 
 - Covers logistical and practical considerations (e.g., "Must be able to attend all clinic
 visits," "No plans to move out of state," "Language proficiency requirements").

**Instructions**:
1. Choose **exactly one** domain that best fits the criterion.
2. Provide your answer in **valid JSON** with the key "domain".
3. The value must be one of the **eight** domains shown above.
4. Do **not** include any other text, explanations, or commentary. 
\end{verbatim}

The user prompt included the description of the eligibility criterion as well as the name of trial it belongs to.

\section{Classification of eligibility criteria: requested data format}
\label{sec:data_format_classification_appendix}
Below is the prompt used to classify eligibility criteria by requested data format:
\begin{verbatim}
You are a data-format classification assistant for clinical trial eligibility criteria.

**Task**:
You will receive exactly one clinical trial eligibility criterion.
Your job is to decide if it is primarily “structured” or “unstructured.”

1. **Structured** 
 - The criterion can be evaluated using discrete or numeric data fields, codes, 
 or straightforward yes/no inputs.
 - Often involves specific numeric cutoffs, ICD codes, or standardized pick-list fields.
 - Example: "Hemoglobin >10 g/dL."

2. **Unstructured** 
 - The criterion involves textual descriptions that must be derived from free-text 
 medical notes or narratives.
 - Typically requires advanced processing or interpretation of written information.
 - Example: "History of chronic fainting spells, as noted in clinician progress notes."

**Instructions**:
1. Choose exactly one data format that best describes the criterion.
2. Provide your answer in valid JSON with the key "requested_data_format".
3. The value must be either "Structured" or "Unstructured".
4. Do not include any other text, explanations, or commentary.
\end{verbatim}

The user prompt included the description of the eligibility criterion as well as the name of trial it belongs to.

\section{Classification of eligibility criteria: temporal constraint}
\label{sec:temporal_classification_appendix}
Below is the prompt used to classify eligibility criteria by temporal constraint.
\begin{verbatim}
You are a temporal-constraint classification assistant for 
clinical trial eligibility criteria.

**Task**:
You will receive exactly one clinical trial eligibility criterion.
Your job is to decide if it explicitly requires a time component (e.g., “within 6 months,” 
“for at least 3 months,” “2 weeks prior,” etc.).

1. **Yes** 
   - The statement explicitly uses time-based expressions (last 6 months, 
   at least 2 years, prior to Visit 1).

2. **No** 
   - The statement does not reference any time window or duration.

**Instructions**:
1. Choose exactly one value that best describes whether a time constraint exists.
2. Provide your answer in valid JSON with the key "temporal_constraint".
3. The value must be either "Yes" or "No".
4. Do not include any other text, explanations, or commentary.
\end{verbatim}

The user prompt included the description of the eligibility criterion as well as the name of trial it belongs to.

\section{Heuristics to deduce time taken by users to review patients}
\label{sec:time_heuristics_appendix}
Our heuristic was as follows:
\begin{enumerate}
    \item We considered only patient-trial pairs where users had reviewed at least two criteria and had also classified the overall patient as either \textit{Eligible for screening} or \textit{Not eligible}.
    \item We measured the time elapsed from the review of the first criterion to the moment when the overall patient was classified.
    \item To account for the initial criterion review, we adjusted this duration by a factor of \(\frac{N}{N - 1}\), where \(N\) is the number of criteria reviewed.
    \item We excluded any reviews lasting less than one minute or more than one hour.
\end{enumerate}

This process yielded \textbf{92 patient-trial reviews}.

\section{Research site type definitions}
\label{sec:research_site_type_defintions_appendix}
\textbf{Independent research site embedded in an outpatient clinic}:
A facility that focuses on clinical research while operating within or alongside a routine outpatient clinic.

\textbf{Independent research site}:
A dedicated clinical research facility that operates separately from general healthcare settings. These sites are solely focused on conducting trials and research without being part of a larger hospital or clinic structure.

\textbf{Services company}:
An organization that provides supportive services for clinical trials rather than conducting them directly. This can include functions such as patient recruitment, data management, regulatory consulting, or trial monitoring.

\textbf{Research center part of a site network}:
A research facility that is one member of a larger network of clinical sites.

\textbf{Outpatient clinic}:
A medical facility where patients receive care without hospital admission.

\textbf{Oncology care center}:
A specialized center dedicated to the treatment and care of cancer patients.

\end{document}